\documentclass[runningheads,a4paper]{llncs}

\usepackage{amsmath}
\usepackage{amssymb}
\usepackage{graphicx}

\usepackage{times}
\usepackage{url}
\urldef{\mailsa}\path|{alfred.hofmann, ursula.barth, ingrid.haas, frank.holzwarth,|
\urldef{\mailsb}\path|anna.kramer, leonie.kunz, christine.reiss, nicole.sator,|
\urldef{\mailsc}\path|erika.siebert-cole, peter.strasser, lncs}@springer.com|

\def\eg{{\em e.g.}}
\def\etal{{\em et al.}}

\def\ie{{\em i.e.}}
\newcommand{\bvec}[1]{\mbox{\boldmath $#1$}}
\urldef{\mails}\path|{k.tsukuda, m.goto}@aist.go.jp|

\begin{document}

\mainmatter  

\title{Taste or Addiction?:\\Using Play Logs to Infer Song Selection Motivation}

\toctitle{Addicted To The Artist: Modeling Music Listening Behavior from Play Logs}

\author{Kosetsu Tsukuda and Masataka Goto}

\institute{National Institute of Advanced Industrial Science and Technology (AIST), Japan\\
\mails\\}

\maketitle

\begin{abstract}
Online music services are increasing in popularity.
They enable us to analyze people's music listening behavior based on play logs.
Although it is known that people listen to music based on topic (\eg, rock or jazz), we assume that when a user is addicted to an artist, s/he chooses the artist's songs regardless of topic.
Based on this assumption, in this paper, we propose a probabilistic model to analyze people's music listening behavior.
Our main contributions are three-fold.
First, to the best of our knowledge, this is the first study modeling music listening behavior by taking into account the influence of addiction to artists.
Second, by using real-world datasets of play logs, we showed the effectiveness of our proposed model.
Third, we carried out qualitative experiments and showed that taking addiction into account enables us to analyze music listening behavior from a new viewpoint in terms of how people listen to music according to the time of day, how an artist's songs are listened to by people, etc.
We also discuss the possibility of applying the analysis results to applications such as artist similarity computation and song recommendation.
\end{abstract}

\section{Introduction}\label{sec:introduction}

Among various leisure activities such as watching movies, reading books, and eating delicious food, listening to music is one of the most important for people~\cite{Rentfrow:2003}.
In terms of the amount of accessible music, the advent of online music services (\eg, Last.fm\footnote{\url{http://www.last.fm}}, Pandora\footnote{\url{http://www.pandora.com}}, and Spotify\footnote{\url{http://www.spotify.com}})
has made it possible for people to access millions of songs on the Internet, and it has become popular to play music using such services rather than physical media like CDs~\cite{Kamalzadeh:2012}.
When users play music online, such services record personal musical play logs that show when users listen to music and what they listen to.

Since personal music play logs have become available, it has become popular to use {\it session} information to analyze and model people's music listening behavior~\cite{Baur:2012,Dias:2013,Park:2011,Zheleva:2010}.
Here, a session is a sequence of logs within a given time frame.
Zheleva \etal~\cite{Zheleva:2010} were the first to model listening behavior using a topic model based on session information.
They revealed that a user tends to choose songs in a session according to the session's specific topic such as rock or jazz.
However, it is not always correct to assume that a user chooses songs according to the session's topic.
For example, after a user buys an artist's album or temporarily falls in love with an artist,
s/he will be {\it addicted} to the artist and repeatedly listen to the artist's songs regardless of topic.

In light of the above, this paper proposes a model that can deal with both a session topic and addiction to artists.
Our proposed model uses the model proposed by Zheleva \etal~\cite{Zheleva:2010} as the starting point.
We present each song-listening instance in terms of the corresponding song artist.
In our model, each user has a distribution over topics that reflects the user's usual taste in music and a distribution over artists that reflects the user's addiction to artists.
In addition, each user has a different ratio between usual taste and addiction, and probabilistically chooses a song in a session based on this ratio.
That is, if a user has a high addiction ratio, s/he will probably choose a song of an artist from his/her artist distribution for addiction.
Modeling people's music listening behavior by considering addiction is worth studying from various viewpoints:
\begin{itemize}
\item
Our model can show topic characteristics (\eg, the rock topic has a high ratio of addiction) and artist characteristics (\eg, most users choose an artist's songs when addicted to that artist).
It is important to understand such characteristics from the social scientific viewpoint.
\item
Our model can also show user characteristics (\eg, a user chooses songs based on addiction in a session).
There are many applications that could use this data such as advertisements and recommendation systems.
For example, if a user chooses songs of an artist based on addiction in a session, it would be useful to recommend songs of that artist;
if s/he chooses songs based on a topic, it would be better to recommend other artists' songs in the same topic.
\end{itemize}

Our main contributions in this paper are as follows.
\begin{itemize}
\item
To the best of our knowledge, this is the first study modeling music listening behavior by considering both the usual taste in music and the addiction to artists.
\item
We quantitatively evaluated our model by using real-world music play logs of two music online services.
Our experimental results show that the model adopting both factors achieves the best results in terms of the perplexity computed by using test data.
\item
We carried out qualitative experiments in terms of user characteristics, artist characteristics, and topic characteristics and show that our model can be used to analyze people's music listening behavior from a new viewpoint.
\end{itemize}

The remainder of this paper is organized as follows.
Section~\ref{sec:related_work} presents related work on analyzing music play logs and on modeling music listening behavior.
Section~\ref{sec:model} describes the model that extends the model by Zheleva \etal~\cite{Zheleva:2010} by considering the addiction phenomenon.
Section~\ref{sec:inference} presents a procedure to infer the parameters.
Section~\ref{sec:quantitative_experiments} and \ref{sec:qualitative_experiments} report on our quantitative and qualitative experiments, respectively.
Finally, Section~\ref{sec:conclusion} concludes this paper.

\section{Related Work}\label{sec:related_work}

\subsection{Analysis of Music Listening Behavior}
Analyzing people's music listening behavior has attracted a lot of attention because
(1) understanding how people listen to music is important from the social scientific viewpoint and
(2) the analysis results can give useful insight into various applications such as music player interfaces and recommender systems.

People's music listening behavior has been analyzed from various viewpoints.
Rentfrow and Gosling~\cite{Rentfrow:2003} carried out a questionnaire-based survey and
revealed the correlations between music preferences and personality, self-views (\eg, wealthy and politically liberal), and cognitive ability (\eg, verbal skills and analytical skills).
Renyolds \etal~\cite{Renyolds:2008} made an online survey and reported that environmental metadata such as the user's activity, weather, and location affect the user's music selection.
Analysis by Berkers~\cite{Berkers:2012} using Last.fm play logs showed the significant differences between male and female in terms of their music genre preferences.
More recently, Lee \etal~\cite{Lee:2016} collected responses from users of commercial cloud music services and reported the criteria for generating playlists: personal preference, mood, genre/style, artists, etc.
Among various factors, time information has received a lot of attention.
Herrera \etal~\cite{Herrera:2010} analyzed play counts from Last.fm and discovered that a non-negligible number of listeners listen to certain artists and genres at specific moments of the day and/or on certain days of the week.
Park and Kahng~\cite{Park:2010} used log data of a commercial online music service in Korea and showed that there existed seasonal and time-of-day effects on users' music preference.
Baur \etal~\cite{Baur:2012} also showed the importance of seasonal aspects, which influence music listening, using play logs from Last.fm.

In spite of the variety of listening behavior analyses, to the best of our knowledge, no work has focused on users' addiction to, for example, songs and artists.
In this work, we deal with this factor and analyze people's music listening behavior from a new perspective.

\subsection{Application Based on Music Listening Logs}
Listening logs have been used for various applications, including the detection of similar artists.
Schedl and Hauger~\cite{Schedl:2012} crawled Twitter\footnote{\url{http://twitter.com/}} for the hash tag \#nowplaying and computed artist similarity using co-occurrence-based methods.
Their experimental results showed that listening logs can be used to derive similarity measures for artists.
Another application is playlist generation.
Liu \etal~\cite{Liu:2010} proposed a playlist generation system informed by time stamps of a user's listening logs in addition to the user's music rating history and audio features such as wave forms.
The most popular application is music recommendation.
Since personal music play logs have become available, it has become popular to use session information to recommend songs.
Park \etal~\cite{Park:2011} proposed Session-based Collaborative Filtering (SSCF), which extends traditional collaborative filtering techniques by using preferred songs in the similar session.
Dias and Fonseca~\cite{Dias:2013} proposed temporal SSCF, where for each session, a feature vector is created consisting of five properties including time of day and song diversity.
The work closest to ours is that of Zheleva \etal~\cite{Zheleva:2010}, who proposed a statistical model to describe patterns of song listening.
They showed that a user tends to choose songs in a session according to the session's specific topic.
We will describe the details of their model in Section~\ref{subsec:session_model}.

Although none of these applications used addiction information, we believe that this information could improve the usefulness of these applications.
We discuss the possibility of using our analysis results to improve these applications in Section~\ref{sec:qualitative_experiments}.

\section{Model}\label{sec:model}

As was mentioned earlier, our model builds on the one proposed by Zheleva \etal~\cite{Zheleva:2010}.
After summarizing the notations used in our model in Section~\ref{subsec:notations},
we first describe the model by Zheleva \etal~\cite{Zheleva:2010} in Section~\ref{subsec:session_model} and then propose our model in Section~\ref{subsec:session_with_addiction_model}.

\subsection{Notations}\label{subsec:notations}

Given a music play log dataset, let $U$ be a set of users in the dataset.
Let $l_{un} = (u, a, t_{un})$ denote the $n$th play log of $u \in U$.
More specifically, user $u$ plays a song of artist $a \in A$ at time $t_{un}$.
Here, $A$ is the set of artists in the dataset.
Without loss of generality, we assume that play logs are sorted in ascending order of their timestamps: $t_{un} < t_{un'}$ for $n < n'$.

To capture user's listening preferences over time, we divide user's play logs into sessions.
Following Zheleva \etal~\cite{Zheleva:2010} and Baur \etal~\cite{Baur:2012}, we use the time gap approach to generate sessions.
If the gap between $t_{un}$ and $t_{un+1}$ is less than 30 minutes, $l_{un}$ and $l_{un+1}$ belong to the same session; otherwise, they belong to different sessions.
Let $S_{ur}$ be the $r$th session of $u$ where $S_{ur}$ consists of one or more of $u$'s logs.
Let $R_{u}$ be the total number of $u$'s sessions; then the set of $u$'s sessions is given by $D_{u} = \{ S_{ur} \}_{r=1}^{R_{u}}$.
Hence, the set of sessions of all users is given by $D = \{ D_{u} \}_{u \in U}$.

\begin{figure}[!t]
\centering
\includegraphics[scale=0.27]{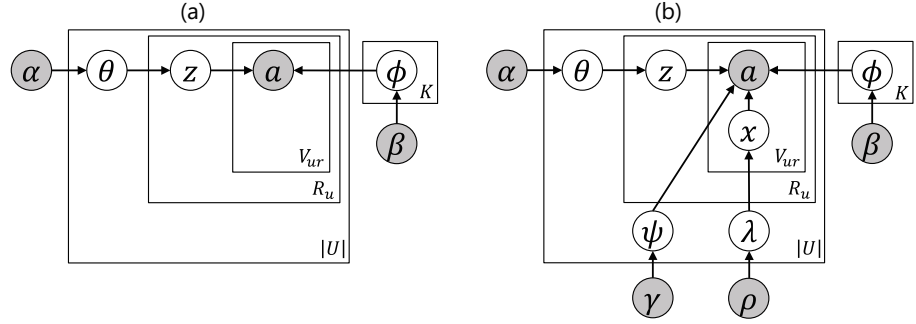}
\caption{Graphical models of (a) session model and (b) session with addiction model.}
\vspace{-1em}
\label{fig:graphical_model}
\end{figure}

\subsection{Session Model}\label{subsec:session_model}

The model proposed by Zheleva \etal~\cite{Zheleva:2010}, which is called the session model, is a probabilistic graphical model based on the Latent Dirichlet Allocation (LDA)~\cite{Airoldi:2008}.
The session model assumes that for each session, there is a latent topic (\eg, rock or love song) that guides the choice of songs in the session.
Figure~\ref{fig:graphical_model}(a) shows the graphical model of the session model, where shaded and unshaded circles represent observed and unobserved variables, respectively.
In the figure, $K$ is the number of topics, $V_{ur}$ is the number of logs in the $r$th session of $u$, $\theta$ is the user-topic distribution, and $\phi$ is the topic-artist distribution.
We assume that $\theta$ and $\phi$ have Dirichlet priors of $\alpha$ and $\beta$, respectively.
The generative process of the session model is as follows:
\vspace{-0.5em}
\begin{itemize}
\item For each topic $k \in \{1, \cdots, K\}$, draw $\phi_{k}$ from $Dirichlet(\beta)$.
\item For each user $u$ in $U$,
	\begin{itemize}
	\item Draw $\theta_{u}$ from $Dirichlet(\alpha)$.
	\item For each session $S_{ur}$ in $D_{u}$,
		\begin{itemize}
		\item Draw a topic $z_{ur}$ from $Categorical(\theta_{u})$.
		\item For each song in $S_{ur}$, observe an artist $a_{urj}$ from $Categorical(\phi_{z_{ur}})$.
		\end{itemize}
	\end{itemize}
\end{itemize}
\vspace{-0.5em}
In the generative process, $a_{urj}$ represents the $j$th song's artist in the $r$th session of $u$.

\subsection{Session with Addiction (SWA) Model}\label{subsec:session_with_addiction_model}

Although Zheleva \etal~\cite{Zheleva:2010} reported the usefulness of generating played songs based on a session's topic,
we hypothesize that users can choose a song independently of topic.
For example, after a user buys an artist's album or temporarily falls in love with an artist, s/he will repeatedly listen to the artist's songs regardless of the topic.
In other words, the user can be {\it addicted} to some artists.
In such an addiction mode, we assume that the user directly chooses a song without going through the topic.

In light of the above, our model takes both session-topic-based and addiction-based choices of songs.
Figure~\ref{fig:graphical_model}(b) shows the graphical model of our proposed model.
Each user has a Bernoulli distribution $\lambda$ that controls the weights of influence for a session topic and addiction.
To be more specific, when user $u$ chooses a song in a session, we assume that the choice is influenced by the session topic with probability $\lambda_{u0}$ $(x=0)$ and
by $u$'s addiction to the artist with probability $\lambda_{u1}$ $(x=1)$, where $\lambda_{u0} + \lambda_{u1} = 1$.
When $x=0$, a song is generated through the same process of the session model,
while when $x=1$, a song is directly generated from a user-artist distribution $\psi$.
The generative process of the SWA model is as follows:
\vspace{-0.5em}
\begin{itemize}
\item For each topic $k \in \{1, \cdots, K\}$, draw $\phi_{k}$ from $Dirichlet(\beta)$.
\item For each user $u$ in $U$,
	\begin{itemize}
	\item Draw $\theta_{u}$ from $Dirichlet(\alpha)$.
	\item Draw $\psi_{u}$ from $Dirichlet(\gamma)$.
	\item Draw $\lambda_{u}$ from $Beta(\rho)$.
	\item For each session $S_{ur}$ in $D_{u}$,
		\begin{itemize}
		\item Draw a topic $z_{ur}$ from $Categorical(\theta_{u})$.
		\item For each song in $S_{ur}$,
			\begin{itemize}
			\item Sample $x$ from $Bernoulli(\lambda_{u})$.
			\item If $x=0$, observe an artist $a_{urj}$ from $Categorical(\phi_{z_{ur}})$.
			\item If $x=1$, observe an artist $a_{urj}$ from $Categorical(\psi_{u})$.
			\end{itemize}
		\end{itemize}
	\end{itemize}
\end{itemize}
\vspace{-0.5em}

\section{Inference}\label{sec:inference}

To learn the parameters of our proposed model, we use collapsed Gibbs sampling~\cite{Griffiths:2004} to obtain samples of hidden variable assignment.
Since we use a Dirichlet prior for $\theta$, $\phi$, and $\psi$ and a Beta prior for $\lambda$, we can analytically calculate the marginalization over the parameters.
The marginalized joint distribution of $D$, latent variables $Z=\{ \{z_{ur}\}_{r=1}^{R_{u}} \}_{u \in U}$, and latent variables $X=\{\{\{x_{urj}\}_{j=1}^{V_{ur}}\}_{r=1}^{R_{u}}\}_{u \in U}$ is computed as follows:
\begin{align}
&P(D,Z,X|\alpha,\beta,\gamma,\rho) \notag \\
&=\iiiint P(D,Z,X|\bvec{\Theta}, \bvec{\Phi},\bvec{\Psi},\bvec{\Lambda}) P(\bvec{\Theta}|\alpha) P(\bvec{\Phi}|\beta) P(\bvec{\Psi}|\gamma) P(\bvec{\Lambda}|\rho) d\bvec{\Theta} d\bvec{\Phi} d\bvec{\Psi} d\bvec{\Lambda}, \label{eq:mjd}
\end{align}
where $\bvec{\Theta} = \{\theta_{u}\}_{u \in U}$, $\bvec{\Phi} = \{ \phi_{k} \}_{k=1}^{K}$, $\bvec{\Psi} = \{ \psi_{u} \}_{u \in U}$, and $\bvec{\Lambda} = \{ \lambda_{u}\}_{u \in U}$.
By integrating out those parameters, we can compute Equation (\ref{eq:mjd}) as follows:
\begin{align}
&P(D,Z,X|\alpha,\beta,\gamma,\rho) \notag \\
&= \left( \frac{{\rm \Gamma} (2\rho)}{{\rm \Gamma}(\rho)^{2}} \right)^{|U|} \prod_{u \in U} \frac{{\rm \Gamma}(\rho + N_{u0}) {\rm \Gamma}(\rho + N_{u1})}{{\rm \Gamma}(2\rho + N_{u})} \left( \frac{{\rm \Gamma}(\gamma |A|)}{{\rm \Gamma}(\gamma)^{|A|}} \right)^{|U|} \prod_{u \in U} \frac{\prod_{a \in A} {\rm \Gamma}(N_{u1a} + \gamma)}{{\rm \Gamma}(N_{u1} + \gamma |A|)} \notag \\
& \times \left( \frac{{\rm \Gamma}(\beta |A|)}{{\rm \Gamma}(\beta)^{|A|}} \right)^{K} \prod_{k = 1}^{K} \frac{\prod_{a \in A} {\rm \Gamma}(N_{ka} + \beta)}{{\rm \Gamma}(N_{k} + \beta |A|)} \left( \frac{{\rm \Gamma}(\alpha K)}{{\rm \Gamma}(\alpha)^{K}} \right)^{|U|} \prod_{u \in U} \frac{\prod_{k=1}^{K} {\rm \Gamma}(R_{uk} + \alpha)}{{\rm \Gamma}(R_{u} + \alpha K)}.
\end{align}
Here, $N_{u0}$ and $N_{u1}$ are the number of $u$'s logs such that $x=0$ and $x=1$, respectively, and $N_{u} = N_{u0} + N_{u1}$.
The term $N_{u1a}$ represents the number of times that user $u$ chooses artist $a$'s song under the condition of $x=1$, and $N_{u1} = \sum_{a \in A} N_{u1a}$.
Furthermore, $N_{k} = \sum_{a \in A} N_{ka}$ where $N_{ka}$ is the number of times artist $a$ is assigned to topic $k$ under the condition of $x=0$.
Finally, $R_{uk}$ is the number of times $u$'s session is assigned to topic $k$, and $R_{u} = \sum_{k=1}^{K} R_{uk}$.

For the Gibbs sampler, given the current state of all but one variable $z_{ur}$, the new latent assignment of $z_{ur}$ is sampled from the following probability:
\begin{align}
&P(z_{ur} = k | D, X, Z_{\setminus ur}, \alpha, \beta, \gamma, \rho) \notag \\
& \propto \frac{R_{uk \setminus ur} + \alpha}{R_{u} -1 + \alpha K} \frac{{\rm \Gamma}(N_{k \setminus ur} + \beta |A|)}{{\rm \Gamma}(N_{k \setminus ur} + N_{ur} + \beta |A|)} \prod_{a \in A} \frac{{\rm \Gamma}(N_{ka \setminus ur} + N_{ura} + \beta)}{{\rm \Gamma}(N_{ka \setminus ur} + \beta)},
\end{align}
where $\setminus ur$ represents the procedure excluding the $r$th session of $u$.
Moreover, $N_{ur}$ and $N_{ura}$ represent the number of logs in $r$th session of $u$ and the number of $a$'s logs in $r$th session of $u$, respectively.

In addition, given the current state of all but one variable $x_{urj}$, the probability at which $x_{urj} = 0$ is computed as follows:
\begin{equation}
P(x_{urj} = 0 | D, X_{\setminus urj}, Z, \alpha, \beta, \gamma, \rho) \propto \frac{\rho + N_{u0 \setminus urj}}{2\rho + N_{u} - 1} \frac{N_{z_{ur}a_{urj} \setminus urj} + \beta}{N_{z_{ur} \setminus urj} + \beta |A|}, \label{eq:degree_of_taste}
\end{equation}
where $\setminus urj$ represents the procedure excluding the $j$th song in the $r$th session of $u$.
Similarly, the probability at which $x_{urj} = 1$ is computed as follows:
\begin{equation}
P(x_{urj} = 1 | D, X_{\setminus urj}, Z, \alpha, \beta, \gamma, \rho) \propto \frac{\rho + N_{u1 \setminus urj}}{2\rho + N_{u} - 1} \frac{N_{u1a_{urj} \setminus urj} + \gamma}{N_{u1 \setminus urj} + \gamma |A|}. \label{eq:degree_of_addiction}
\end{equation}

Finally, we can make the point estimates of the integrated out parameters as follows:
\begin{equation}
\theta_{uk} = \frac{R_{uk} + \alpha}{R_{u} + \alpha K}, \ \ \ \phi_{ka} = \frac{N_{ka} + \beta}{N_{k} + \beta |A|}, \ \ \ \psi_{ua} = \frac{N_{u1a} + \gamma}{N_{u1} + \gamma |A|}. \label{eq:estimate_theta_phi_psi}
\end{equation}
\begin{equation}
\lambda_{u0} = \frac{N_{u0} + \rho}{N_{u} + 2\rho},\ \ \ \lambda_{u1} = \frac{N_{u1} + \rho}{N_{u} + 2\rho}, \label{eq:estimate_lambda}
\end{equation}
where remind that $\lambda_{u0}$ and $\lambda_{u1}$ represent the ratio of usual taste in music and addiction when $u$ chooses songs, respectively.

\section{Quantitative Experiments}\label{sec:quantitative_experiments}

In this section, we answer the following research question based on our quantitative experimental results:
is adopting two factors, which are users' daily taste in music and addiction to artists, effective to model music listening behavior?

\subsection{Dataset}\label{subsec:dataset}

To examine the effectiveness of the proposed model, we constructed two datasets.
The first one is created from music play logs on a music download service in Japan.
On the service, users can buy a single song and an album and listen to them.
For this evaluation, we obtained 10 weeks of log data between 1/1/2016 and 10/3/2016.
We call this dataset JPD.
The second one consists of logs on Last.fm.
To guarantee the repeatability, we used a publicly available music play log data on Last.fm provided by Schedl~\cite{Schedl:2016}.
Similar with JPD, we extracted 10 weeks of log data between 1/1/2013 and 11/3/2013; we call the dataset LFMD.

From the 10 weeks of data of JPD, we created two pairs of training and test datasets as follows.
In the first/second dataset, the training dataset consists of logs of the first four/eight weeks and the test dataset consists of the next two weeks.
For each dataset, we excluded artists whose songs were played by $\leq 3$ users and created session data as described in Section~\ref{subsec:notations}.
Let the first and second dataset be 4WJPD (4W means four weeks) and 8WJPD, respectively.
As for LFMD, we also created two pairs of training and test datasets 4WLFMD and 8WLFMD in the same manner as we created the 4WJPD and 8WJPD datasets.
Table~\ref{tbl:dataset_statistics} shows the statistics of the four datasets.

\begin{table}[t!]
\begin{center}
\caption{Statistics of our datasets}
\scalebox{1.0}{
\begin{tabular}{l|r|r|r|r}
\hline
                & 4WJPD & 8WJPD & 4WLFMD & 8WLFMD \\ \hline
Number of users                                & 7,230      &  13,986     & 2,501     & 2,850 \\
Number of artists                              & 3,441       &  6,431       & 7,899     &  12,360  \\
Number of logs in training data        & 141,381   &  331,437   & 400,410 & 872,614 \\
Number of sessions in training data & 35,780     &  82,427     & 50,106    & 106,840 \\
Number of logs in test data              & 48,837    &  57,126     & 179,983   & 201,966 \\
Number of sessions in test data       & 11,767    &  13,516     & 23,167     & 24,958 \\
\hline
\end{tabular}
}
\vspace{-2em}
\label{tbl:dataset_statistics}
\end{center}
\end{table}

\subsection{Settings}\label{sebsec:settings}

In terms of hyperparameters, in line with other topic modeling work, we set $\alpha = \frac{1}{K}$ and $\beta = \frac{50}{|A|}$ in the session model and the session with addiction (SWA) model.
In addition, in the SWA model, we set $\gamma = \frac{50}{|A|}$ and $\rho = 0.5$.

To compare the performance of the session model and  the SWA model, we use the perplexities of the two models.
Perplexity is a widely used measure to compare the performance of statistical models~\cite{Airoldi:2008} and the lower value represents the better performance.
The perplexity of each model on the test data is given by:
\begin{equation}
{\it perplexity}(D_{\it test}) = \exp \left( - \frac{\sum_{u \in U} \sum_{r=1}^{R_{u}^{{\it test}}} \sum_{j=1}^{V_{ur}^{{\it test}}} p(a_{urj})}{\sum_{u \in U} \sum_{r=1}^{R_{u}^{{\it test}}} |V_{ur}^{{\it test}}|} \right),
\end{equation}
where $R_{u}^{{\it test}}$ and $V_{ur}^{{\it test}}$ represent the number of $u$'s sessions and the number of logs in $r$th session of $u$ in the test data, respectively.
The $p(a_{urj})$ is computed based on the estimated parameters obtained by Equation (\ref{eq:estimate_theta_phi_psi}) and (\ref{eq:estimate_lambda}) as follows:
\begin{equation}
p(a_{urj}) = \lambda_{u0} \sum_{k=1}^{K} \theta_{uk} \phi_{ka_{urj}} + \lambda_{u1} \psi_{ua_{urj}}.
\end{equation}
In terms of the number of topics, we compute the perplexity for $K=$ 5, 10, 20, 30, 40, 50, 100, 200, and 300.

\subsection{Results}\label{subsec:results}

\begin{figure}[!t]
\centering
\includegraphics[scale=0.45]{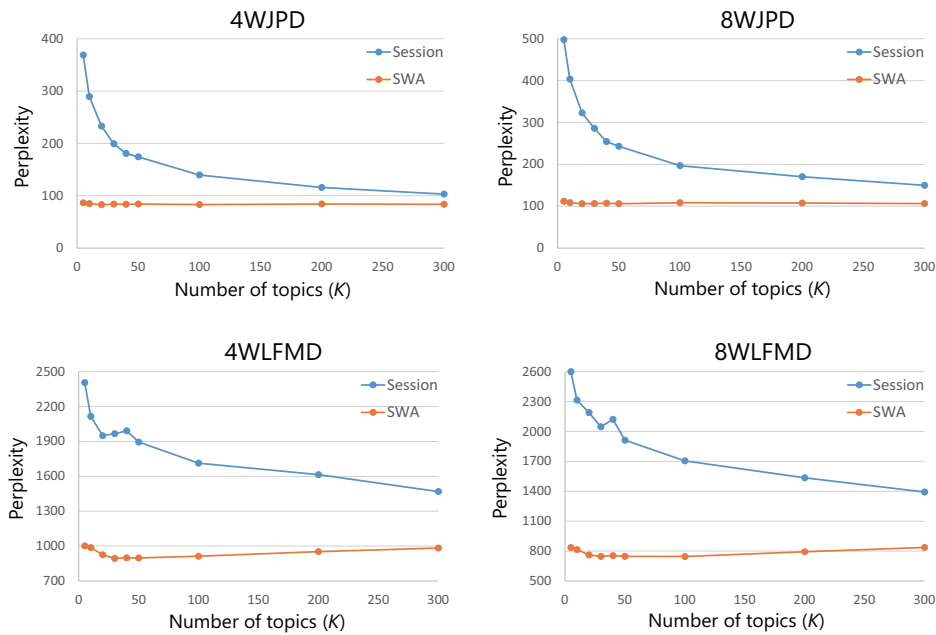}
\vspace{-1em}
\caption{Perplexity for 4WJPD, 8WJPD, 4WLFMD, and 8WLFMD.}
\vspace{-1em}
\label{fig:perplexity}
\end{figure}

Figure~\ref{fig:perplexity} shows the perplexity for each dataset.
In any dataset, regardless of the amount of training data and the number of topics, the SWA model outperformed the session model.
If we set the number of topics to be larger than 300, the session model might outperform the SWA model;
but we set the maximum value of $K$ to 300 for the following two reasons.
The first reason is due to the expended hours for the learning process.
For example, when the session model learns parameters for $K=300$ using 8WJPD, it takes 9.8 times longer than the SWA model does for $K=30$ using 8WJPD (1,713 minutes for the session model and 175 minutes for the SWA model).
In data analysis, the expended hours is an important factor; if it takes a long time to learn the parameters for a model, the model is inappropriate for data analysis.
The second reason is due to the understandability of topics.
When the number of topics becomes too large, it is difficult to understand the difference between topics because there are many similar topics.
As we will show in Section~\ref{subsec:topic_characteristics}, analyzing the characteristics of each topic is useful to understand people's music listening behavior.
Hence, it is undesirable to set $K$ to a large value.
For these reasons, we conclude that the SWA model is a better model than the session model.

\section{Qualitative Experiments}\label{sec:qualitative_experiments}

In this section, we report on the qualitative analysis results in terms of user characteristics, artist characteristics, and topic characteristics.
Due to the space limitation, we only show the results for the training data of 8WJPD with $K=30$.
We not only analyze people's music listening behavior but discuss how we can apply the analysis results.

\subsection{User Characteristics}\label{subsec:user_characteristics}

\begin{figure}[!t]
\centering
\includegraphics[scale=0.43]{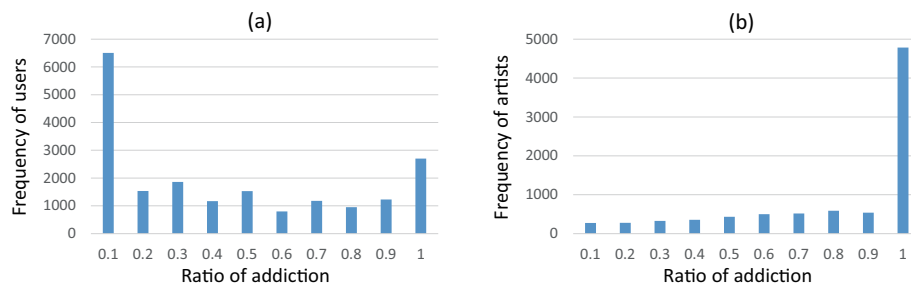}
\vspace{-2em}
\caption{Histogram based on ratio of addiction among (a) users and (b) artists.}
\vspace{-0.5em}
\label{fig:user_and_artist_addiction}
\end{figure}

As we mentioned in Section~\ref{subsec:session_with_addiction_model}, each user has a parameter $\lambda$ that controls the degree of usual taste in music and addiction when s/he chooses songs.
Given a user $u$, we can obtain the ratio of these two factors from Equation (\ref{eq:estimate_lambda}), where $\lambda_{u0} + \lambda_{u1} = 1$.
Figure~\ref{fig:user_and_artist_addiction} (a) shows a histogram based on the degree of addiction.
Although most people put a high priority on their usual taste in music (ratio $\le$ 0.1), the second highest histogram peak is for those who put the greatest weight on addiction to artists (ratio $>$ 0.9).
The result where so many users lie somewhere between these two extremes of behavior further indicates the usefulness of considering the addiction mode in music listening behavior.

\begin{figure}[!t]
\centering
\includegraphics[scale=0.4]{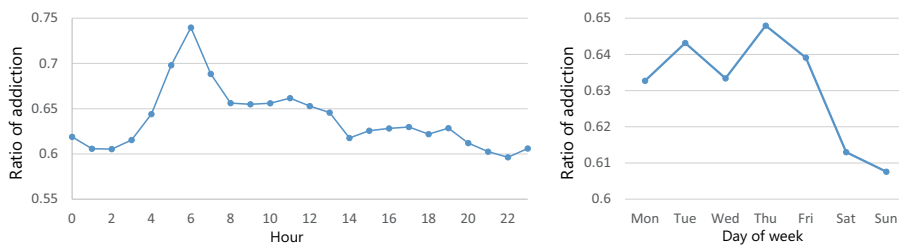}
\vspace{-1em}
\caption{Time-dependent ratio of addiction: per-hour analysis result and per-weekday analysis result.}
\vspace{-1.5em}
\label{fig:time_addiction}
\end{figure}

By using the posterior distribution of latent variables in Equation (\ref{eq:degree_of_taste}) and (\ref{eq:degree_of_addiction}), we can analyze the relationship between the degree of addiction and the time.
We first analyzed the transition of the degree of addiction on a per-hour basis.
For example, to analyze the degree between 9:00:00 and 9:59:59, we collected all play logs during the time period in the training data.
By summing $p(x=0)$ of all logs, we can obtain the strength of usual taste in music during the time period.
Similarly, by summing $p(x=1)$ of all logs, we can obtain the strength of addiction during the time period.
Finally, we normalize their sum to 1 so that we can see the ratio of the degree of the two factors.
The left line chart in Figure~\ref{fig:time_addiction} shows the results.
It can be observed that the degree of addiction is high in the early morning (\ie, at 5, 6, and 7 am), while it is low at night (\ie, at 9, 10, and 11 pm).
We can estimate that people tend to be short on time in the morning, and as a result, they listen to a specific artist's songs rather than choosing various songs according to a topic.
On the other hand, at night, people have time to spare and tend to listen to various artists' songs by choosing from a topic.
These results indicate that the transition of the degree of addiction on a per-hour basis enables us to analyze people's music listening behavior from a new viewpoint.
In addition, we propose applying the knowledge to music recommendation.
For example, it would be more appropriate to recommend unknown songs to the user at night rather than in the morning because s/he would have time to try listening to new songs.

In the same manner as the above analysis, we also analyzed the transition of the degree of addiction on a day of the week basis.
The right line chart in Figure~\ref{fig:time_addiction} shows the result.
It can be observed that the degree of addiction is high on weekdays, while it is low on weekends.
We can also estimate that the degree of addiction is high on weekdays because people are busy working on weekdays, while the degree is low on weekends because people have more time.
These results would also be useful to recommend music.

\vspace{-0.5em}
\subsection{Artist Characteristics}\label{subsec:artist_characteristics}

In the same way as Section~\ref{subsec:user_characteristics}, given an artist, by summing $p(x=0)$ and $p(x=1)$ of all the artist's logs,
we can obtain the strength of usual taste and addiction during the time period, respectively.
Then their sum is normalized to 1 to compute the ratio of each factor of the artist.
Figure~\ref{fig:user_and_artist_addiction} (b) shows a histogram based on the degree of addiction.
It can be observed that most artists have a high degree of addiction.
From these results, we can estimate whether the artist's songs are repeatedly played by users who are enthusiastic admirers of the artist or by various users who listen to the artist's songs with other artists' songs.
In addition, we believe that the results could be used as one of the features to compute the similarity between artists by assuming that similar artists have similar degrees of addiction.

\vspace{-0.5em}
\subsection{Topic Characteristics}\label{subsec:topic_characteristics}

\begin{figure}[!t]
\centering
\includegraphics[scale=0.55]{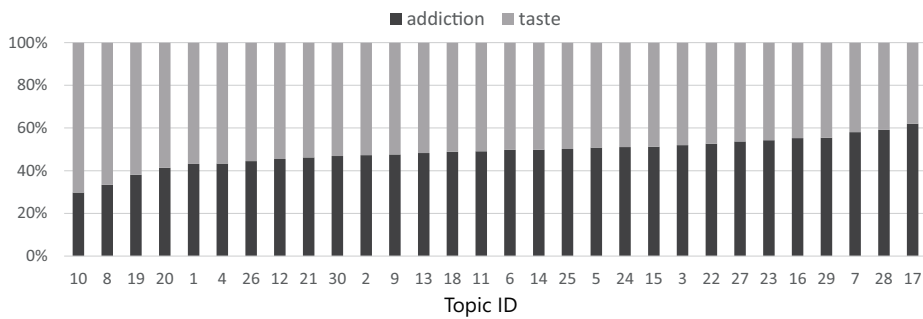}
\vspace{-2em}
\caption{Ratio of taste in music and addiction for each topic.}
\vspace{-1em}
\label{fig:topic_addiction}
\end{figure}

Finally, we show that our model can also be used for topic analysis.
Given a topic $k$, we collected representative artists in the category.
To be more specific, the top 20 artists in terms of $\phi_{k}$ were extracted.
For each of the 20 artists, we collected all logs in the training data and computed the ratio of the degree of taste in music and addiction as described in Section~\ref{subsec:user_characteristics}.
We then computed the average values of each degree over 20 artists and normalized their sum to 1.
Figure~\ref{fig:topic_addiction} shows the ratio of 30 topics, where topics are sorted in ascending order of addiction ratio.
As can be seen, the ratio between two factors is largely different from one topic to another: the addiction ratio ranged from 0.297 (10th topic) to 0.620 (17th topic).
As for the low addiction topics, the 10th topic has the lowest value of 0.297.
This topic is related to songs created by using VOCALOID~\cite{Kenmochi:2007}, which is popular singing synthesizer software in Japan.
The 8th topic has the second lowest value of 0.334 and its topic is related to anime songs.
From these results, we can estimate that when people listen to music related to popular culture, they tend to listen to various artists' songs in the topic.
As for the high addiction topic, the 17th topic, which is related to Western artists, and the 28th topic, which is related to old Japanese artists, have the highest values of 0.620 and 0.592, respectively.
These results indicate the possibility of applying the knowledge to playlist generation.
In topics with a high addiction degree, it would be useful to generate a playlist that consists of songs of a specific artist;
while in topics with a low addiction degree, it would be useful to generate a playlist that consists of various artists' songs.

\vspace{-0.5em}
\section{Conclusion}\label{sec:conclusion}
In this paper we proposed a probabilistic model for analyzing people's music listening behavior.
The model incorporates the user's usual taste in music and addiction to artists.
Our experimental results using real-world music play logs showed that our model outperformed an existing model that considers only the user's taste in terms of perplexity.
In our qualitative experiments, we showed the usefulness of our model in various aspects: time-dependent play log analysis (\eg, the degree of addiction is high in the early morning and on weekdays),
topic-dependent play log analysis (\eg, the degree of addiction is low in an anime song topic), etc.

For future work, we are interested in applying the knowledge obtained from log analysis to applications such as artist similarity computation and song recommendation as discussed in Section~\ref{sec:qualitative_experiments}.
We are also interested in extending our model by considering the time transition of addiction.
For example, a user who is addicted to some artists in summer may be addicted to largely different artists in autumn.
Considering such time dependency by using the topic tracking model~\cite{Iwata:2009} is one possible direction to take to extend our model.

\vspace{-1em}
\section*{Acknowledgements}
\vspace{-0.5em}
This work was supported in part by ACCEL, JST.

\vspace{-1em}
\bibliographystyle{abbrv}
\bibliography{PAKDD2017}

\end{document}